\documentclass[11pt,a4paper, oneside]{article}
\usepackage{amsmath}
\usepackage{multirow}
\usepackage{epsfig,subfigure,float}

\begin{document}

\title{Three Cases of Connectivity and Global Information Transfer in Robot Swarms}
\author{Serge Kernbach\\
\textit{\small Institute of Parallel and Distributed Systems, University of Stuttgart,} \\
\textit{\small Universit{\"a}tsstr.~38, D-70569 Stuttgart, Germany}\\
\emph{\small Serge.Kernbach@ipvs.uni-stuttgart.de}}
\date{}

\maketitle

\begin{abstract}
In this work we consider three different cases of robot-robot interactions and resulting global information transfer in robot swarms. These mechanisms define cooperative properties of the system and can be used for designing collective behavior. These three cases are demonstrated and discussed based on experiments in a swarm of microrobots "Jasmine".
\end{abstract}

\section{Introduction}

Collective (swarm) systems behave like one organism, therefore they are so fascinating: there are a large number of elements (agents), but no visible coordinator. Biologists see in the swarm behavior a superior of natural systems~\cite{Bonabeau99}, engineers see an example of how a huge complexity can be very elegantly treated~\cite{Eberhart01}. Social researches mention a swarm as a perfectly coordinated society, physicians look for "swarm rules"~\cite{Luna00}, community of collective AI~\cite{Kornienko_S05b} investigates principles underlying swarm intelligence. Lately swarm systems became a research object within robotic domain as the swarm robotics~\cite{Sahin04}.

Developing communication mechanisms for microrobotic swarms, we encounter a few problems of technological and methodological character. First of all, robots have limited communication radius. This allows avoiding the problem of
communication overflow in large-scale swarms (100+ robots), however creates the problem of propagating the relevant information over the swarm. This information concerns an environmental conditions, e.g., energy resources or dangers, and swarm-internal states such as common behavioral goals. We refer to this mechanism as a global information transfer by using local robot-robot interactions. Robots are limited in hardware for using algorithms and protocols known in the domain of distributed systems~\cite{Coulouris01}, e.g. rule-based coordination~\cite{Durfee88}, token exchange~\cite{Scerri05} or cooperative planing~\cite{Weiss99}.

The amount of information, which needs to be globally transferred, should be minimized to reduce communication overhead, the consumed energy, microcontroller running time and used memory. However, this should be also large enough to guarantee an appropriate collective reaction. In this work we demonstrate three different cases of information transfer in a swarm and point to the underlying issues of collective connectivity and cooperativity among the robots. It is known, that amount of information and its usage in collective systems is related to common knowledge~\cite{Halpern90}. Here we discuss different aspects of common knowledge in connection to swarm connectivity. Generally, this paper summarizes experience, collected within the open-hardware and open-source \emph{swarmrobot.org} project, towards swarm communication in terms of hardware, software, protocols and robot behavior in the swarm of microrobots "Jasmine".

The rest of the paper is organized in the following way. Sec.~\ref{sec:platform} describes briefly the platform. Sec.~\ref{sec:infTransfer} is devoted to global information transfer and connectivity. Finally, Sec.~\ref{sec:conclusion} concludes this work.

\section{Swarm Robot Platform}
\label{sec:platform}

\begin{figure}[ht]
\begin{center}
\subfigure[]{\includegraphics[width=0.49\textwidth]{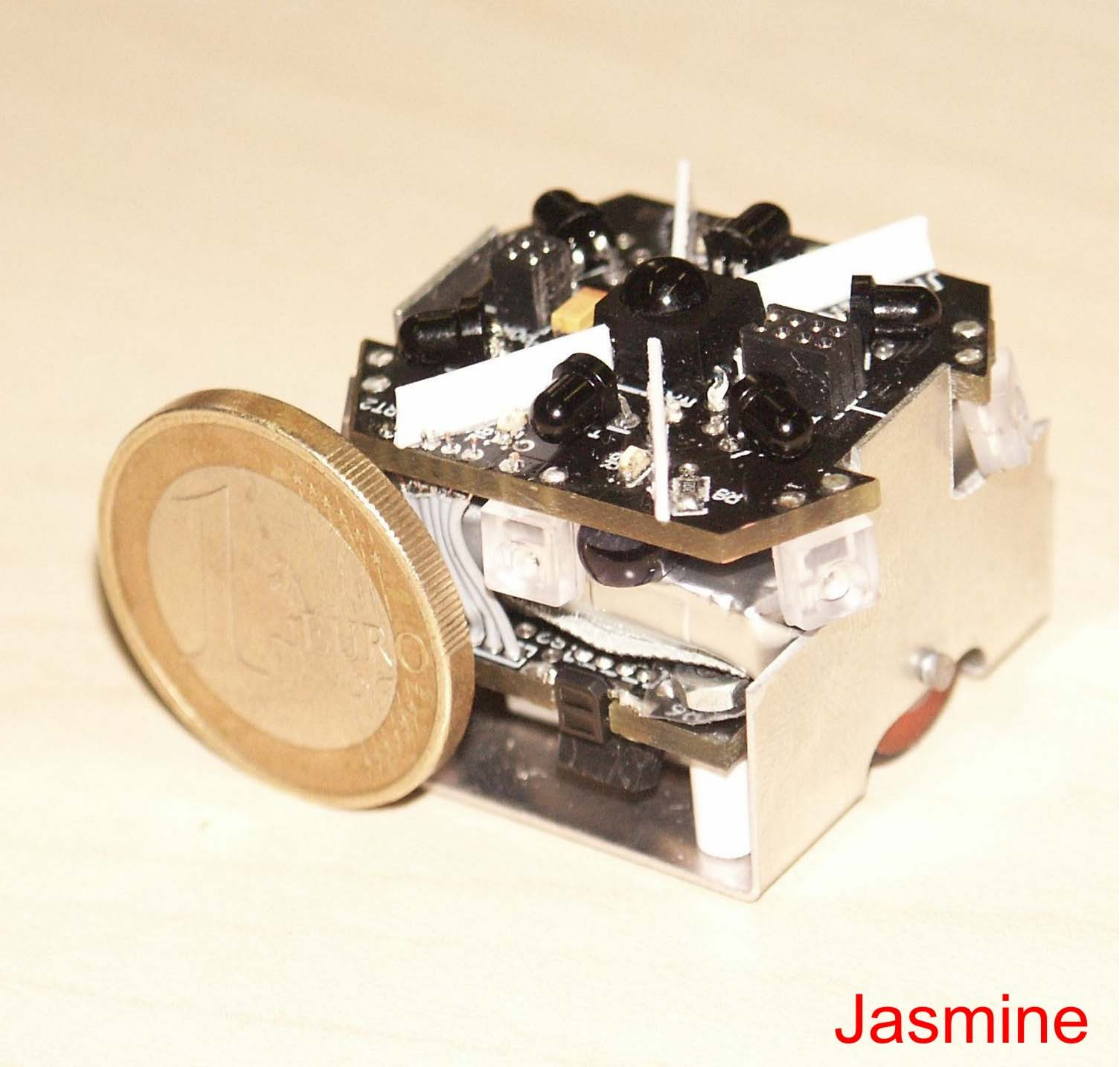}}
\subfigure[]{\includegraphics[width=0.49\textwidth]{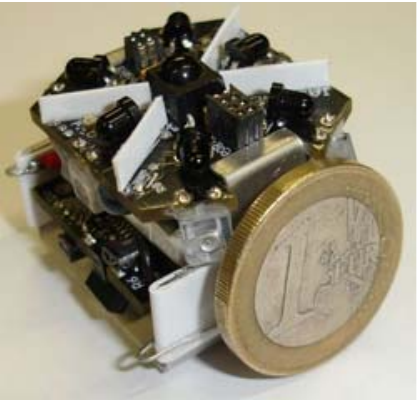}}
\caption{\small \textbf{(a)} The third  and \textbf{(b)} third + versions of the robot "Jasmine".
\label{fig:Jasmine} }
\end{center}
\end{figure}

"Jasmine" Fig.~\ref{fig:Jasmine} is a public open-hardware and open-software development at \textbf{www.swarmrobot.org}, having a goal of creating a simple and cost-effective micro-robot, that could be easy reproduced without special equipment.
The micro-robot is 26$\times$26$\times$20mm (30$\times$30$\times$20mm for III+ version), uses two Atmel AVR Mega micro-controllers: Atmel Mega88 (motor control, odometry, touch, color and internal energy sensing) and Mega168 (communication, sensing, perception, remote control and user defined tasks). Both micro-controllers communicate through high-speed two-wired TWI (I2C) interface. It has in total 24kb flash for program code, 2kb RAM for data and 1kb energy-violent EEPROM for saving working data. Development of communication equipment and general embodiment issues~\cite{Kornienko_S05e} were addressed in \cite{KornienkoS05d}.

The robot uses two DC motors with internal gears, two differentially driven wheels on one axis with a geared motor-wheels coupling. Encoder-less odometrical system normalizes a motion of the robot (the robot is able to move straight forwards and backwards), and estimates the gone distance with accuracy of about 6\% and rotation angle - of 11\%. Jasmine III uses 3V power supply (from 3,7V Li-Po accumulator) with internal IC-stabilization of voltage. Power consumption during a motion is about 200mA, in stand - ~6mA, in stand-by mode less 1 mA. The time of autonomous work is of 1-2 hours. The robot is also capable of autonomous docking and recharging, so that a real time of experiments is in fact unlimited.

The programming of the robot uses C language with open-source \emph{gcc} compiler, there is a complete BIOS system that supports all low-level functions. For a quick implementation of swarm behavior there is a developed jasmine-SDK system, that includes an operating system and high-level functions based on final state automata. See \emph{www.swarmrobot.org} or e.g.~\cite{Kornienko_S05e},~\cite{KornienkoS05d},~\cite{Kornienko_S05d} for details of construction and programming.

\section{Information transfer in a swarm}
\label{sec:infTransfer}

Communication and coordination are well-known issues in collective systems~\cite{Jennings96}. However, these issues in swarm differ from large systems because of hardware limitations and thus include not only technological or algorithmic considerations, but also specific behavior-based mechanisms. The first problem is a routing of information packages in a swarm. We can assume that in the routed package-based communication each package consists of a header with IDs of sender and receiver, routing information and the package content. The package ID is coded by 10 bits, IDs of sender/receiver by 12 bits (6 bit each), so the header is of 22 bits (+1 parity bit), the package content is only of 8 bits. For recording the package history each robot needs about 900 bytes RAM only for routing 300-600 packages within a few minutes (for max $N=50$ robots, each sends max. 1-2 messages each 10 sec, propagation time of 1 min.). The robot has only 1Kb RAM on board. After experiments we came at the conclusion, that pure package routing is not suitable for propagating information through a swarm (however package-based communication is used for local communication between neighbor robots).

The second problem of swarm communication is so-called clusterization. This phenomenon appears when robots fall to groups so that any communication between groups is broken down, see Fig.~\ref{fig:comm3}. When the detached group is small, these robots usually lose "orientation" and are "lost" for swarm. When the group consists of 1/3-1/2 of all robots, this group starts a "parallel" process: building a communication street or looking for resources. When the number of robots is not sufficient in both groups, all robots fail in their collective activities. The strategies for solving both problems consist in specific information exchange mechanisms -- robots receive some message, change it and send further as their own messages. This approach is similar to the infection dynamics. Before we discuss these mechanisms, we have to mention the issue of "common knowledge".

Communication in collective systems is closely related to the notion of "common knowledge", which is well-known in the domain of distributed systems, see e.g.~\cite{Halpern90}. The "common knowledge" describes the degree of collective awareness in a swarm, how the robots are informed about the global and particular states of other robots. The more cooperatively the swarm should operate, the larger "common knowledge" it needs. Therefore in the domain of distributed systems different degrees of "common knowledge" are distinguished (e.g. "all know it", "I know that you know" and so on). The "lowest" degree of common knowledge can easily be achieved (e.g. "I know something about neighbors"), whereas the "highest" degree is very "expensive" from the viewpoint of the resources, required to achieve it. It means that for the cooperative behavior we pay the price of communication effort, computational resources and, finally, the running time and energy. The more intelligent swarm behavior is required, the more advanced cooperation (and so communication and computation) should be involved. This conclusion can be done more generally for distributed economic agents~\cite{Malone87}, planning agents~\cite{Kornienko_S03A}, different  evolutionary~\cite{Schlachter08}, \cite{Kernbach08online} and reconfigurable~\cite{Kernbach08Permis}, \cite{Kernbach08_2} approaches. To some extent, adaptive~\cite{kernbach09adaptive} and artificial~\cite{Kernbach08} self-organization also follows this rule.

\begin {table}[htp]
\fontsize {9} {10} \selectfont
\centering
\begin{tabular}[t]{
p{0.5cm}@{\extracolsep{4mm}}
p{2.4cm}@{\extracolsep{4mm}}
p{2.2cm}@{\extracolsep{4mm}}
p{2.6cm}@{\extracolsep{4mm}}
p{2.6cm}@{\extracolsep{4mm}} } \hline \noalign{\smallskip}
N  & Swarm Capabilities                            & \multicolumn{3}{c}{Degree of cooperativity}
\\\cline{3-5} \noalign{\smallskip}
   &                                               &  {\small \it \bf Basic }         & {\small \it \bf Averaged}      & {\small \it \bf Extended } \\
   &                                               &  {\small \it indirect, }         & {\small \it non-individual}    & {\small \it "individual"} \\
   &                                               &  {\small \it stigmergy}          & {\small \it in groups}         & {\small \it "to individual"} \\ \hline
   & \textbf{ST}                                   &                                  & & \\
1  & Collective movement              & non-coordinated     &  coordinated       & synchronized   \\[3mm]
2  & Building spatial structures,~\cite{Kornienko_S04b} &  "optimizing" patterns & regular patterns (grid, circle, etc.)&  irregular patterns \\[2mm]\hline
   & \textbf{IB}                                   &                                  & & \\
3  & Building informational structures,~\cite{Fu05} & ---    & propagation of information   & "swarm network"\\[2mm]
4  & Collective decision making,~\cite{Kornienko_OS01} & environment--based & agreements techniques & network of multiple decisions \\[2mm]
5  & Collective information processing,~\cite{Geider06}, ~\cite{Fu05} & dynamical systems, \cite{Levi99}, \cite{Kornienko_S06b} & simple distributed processing & distributed processing\\[2mm]
6  & Collective perception \& recognition,~\cite{Kornienko_S05a}  & ---  & collective classification & collective recognition\\[2mm]\hline
   & \textbf{FB}                                   &                                  & & \\
7  & Building functional structures,~\cite{Kornienko_S04a} & --- & dynamic sequences of activities & adaptive functional behavior~\cite{Kernbach10IntSys}\\[2mm]
8  & Collective tasks decomposition and allocation~\cite{Warraich05} & --- & simple decomposition and allocation & dynamic decomposition and allocation\\[2mm]
9  & Collective planning,~\cite{Jimenez05} & --- & --- & simple planning based on tasks decomposition\\[2mm]
10 & Group-based specialization of behavior & --- & ---  & simple functional clusterization\\\hline
\end{tabular}
\caption{\small Swarm capabilities in dependence on the degree of cooperativity, ST -- spatio-temporal, IB -- information-based, FB -- function-based type of swarm behavior. \label{tab_swarm_cap}}
\end{table}

To exemplify this, we collect in Table~\ref{tab_swarm_cap} some "swarm activities" (most of them are experimentally implemented and tested on "Jasmine" robots, see references in the table), which microrobots can collectively perform. These activities are divided into three clusters (spatio-temporal, informational and functional) and into three cases of different cooperativity: stigmergy-like, group-like, and "individual-to-individual". As followed from this table, several activities, especially of functional type, cannot be performed when the coordination level is low. To express cooperativity and, finally, different degrees of information transfer in a swarm, we use the notion of \emph{collective connectivity}. The connectivity means the mechanism that makes a robot aware about other robots and their intensions, and can be \emph{local}, \emph{global} or \emph{feedback-based}. We distinguish between these mechanisms and the \emph{degree of connectivity} $k$, which is defined as the number of information connections between a robot and its $k$ neighbors~\cite{Kernbach11-2-HCR}.

\subsection{Local Connectivity}

\begin{figure}[ht]
\centering
\subfigure[\label{fig:comm11}]{\includegraphics[width=0.49\textwidth]{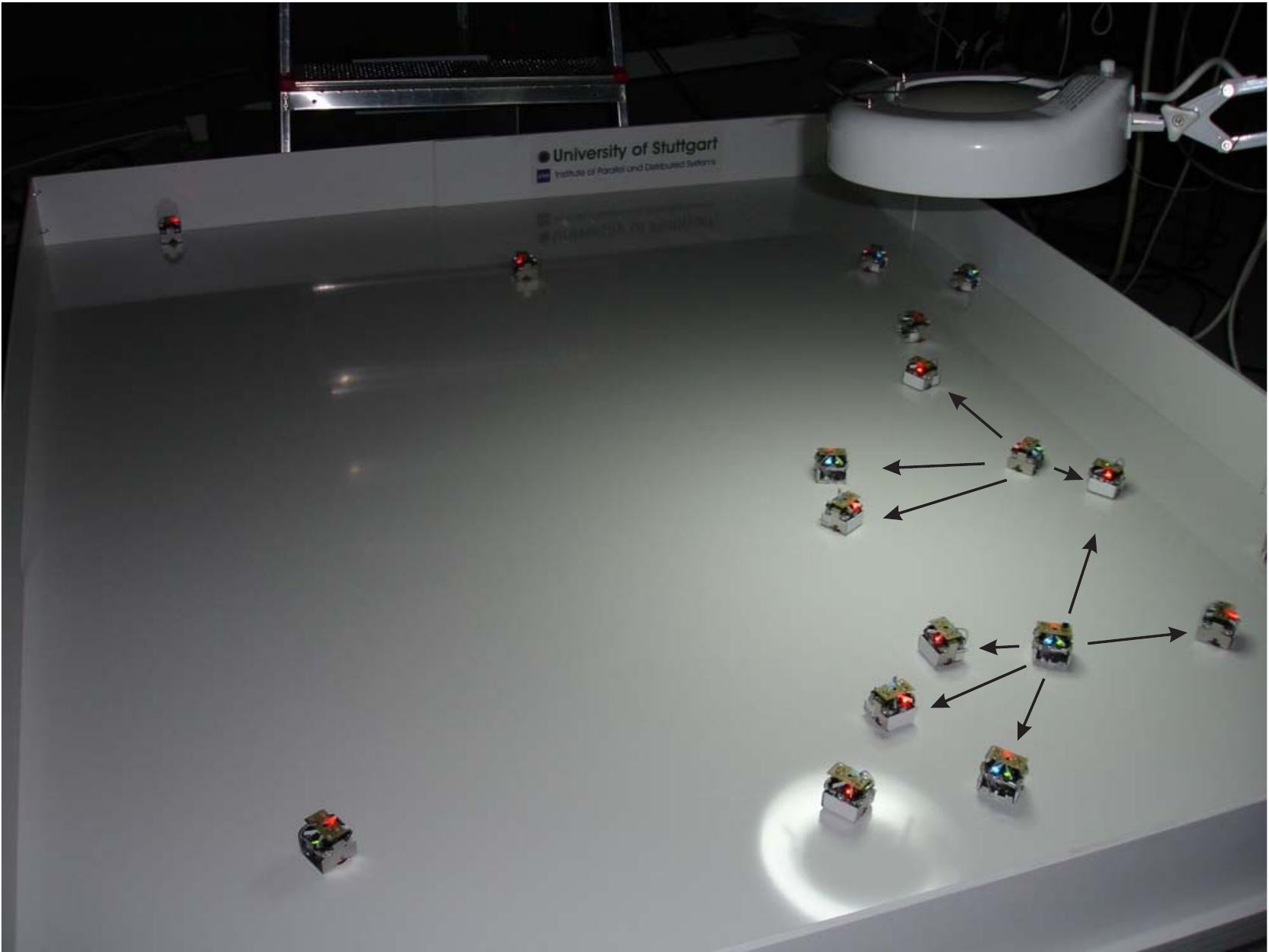}}~
\subfigure[\label{fig:comm12}]{\includegraphics[width=0.49\textwidth]{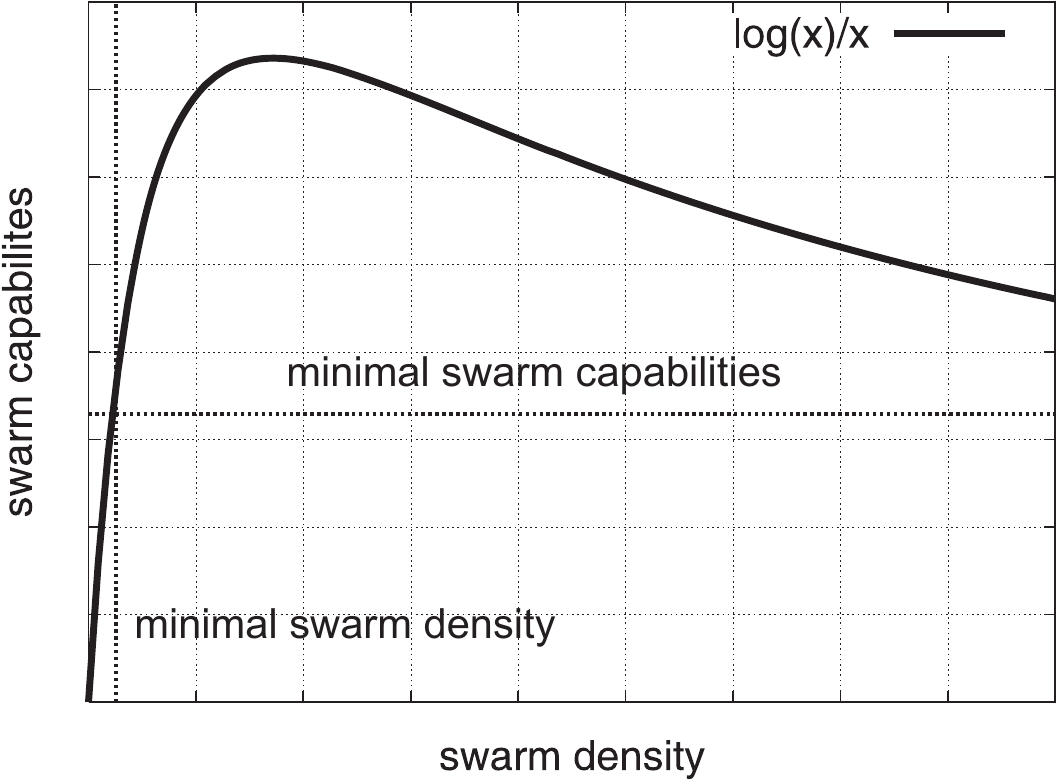}}
\caption{\small \textbf{(a)} Local collective connectivity in experiments with aggregations. Arrows show proximity sensing of two robots. There is no communication in this experiment; \textbf{(b)} Qualitative dependency between swarm capabilities and principal factor for the case of local connectivity.
\label{fig:comm1} }
\end{figure}

Those swarm approaches that originate from biological or physical systems, use mostly the \emph{local connectivity}. Each robot is aware only about its local neighborhood. It is done by means of proximity sensing, by a simple robot-robot communication without messages propagation or by any other local sensors/actuators. In the nature a similar mechanism of indirect communication through environmental changes names stigmergy~\cite{Bonabeau99}. Particular robot does not receive any feedback about its own intensions, i.e. {\it collective behavior is regulated through physical constraints imposed on a swarm and swarm capabilities are primarily defined by kinetic parameters of a swarm~\cite{Kernbach09Nep} and individual sensor/actuator features of robots}. To exemplify this case, we refer to experiments with aggregation around the low-gradient light, see Fig.~\ref{fig:comm1}, see for details~\cite{Kornienko_S06}.

In this experiments, robots are equipped with one light sensor, so  that they cannot find any gradient in the light. The idea of the algorithm is that a robot, when encountering another robot, should stop and wait some time. In this case, the brighter the light is, the longer robots stay and the more robots will be collected under the lamp. When there is already some number of robots under a lamp, they are blocked by new coming robots so that we observe a growing cluster. When there are only a few robots, no aggregation is observed, i.e. this is a typical collective behavior. When the light is moved in another position, robots, after some time, follow the light. We see from this example, that collective aggregation behavior appears without global propagation of information, however due to local physical interactions (collisions). The parameters of collective aggregation are defined by collision avoiding behavior.

The swarm capabilities in this case are defined by kinetic parameters, in particular by the swarm density, see Fig.~\ref{fig:comm1}(b). Increasing the number of robots leads rapidly to growing collective capabilities. However, there is some minimal threshold imposed on the number of robots (for experiments in ~\cite{Kornienko_S06} it is 9 robots), where no collective behavior is observed. Moreover, there is the maximum, after that the swarm capabilities are merely decreasing. Based on works~\cite{Kernbach09Nep},~\cite{Kernbach11-3-HCR}, we estimate a character of this curve as $log(N)/N$. Since there is no propagation of information in this case, the routing and clusterization problems are avoided.

\subsection{Global Connectivity}

As demonstrated by different experiments, the approaches with the local connectivity produce stable and scalable, but relatively simple and mostly only \emph{mono-functional} collective behavior.
\begin{figure}[ht]
\centering
\subfigure[\label{fig:comm21}]{\includegraphics[width=0.49\textwidth]{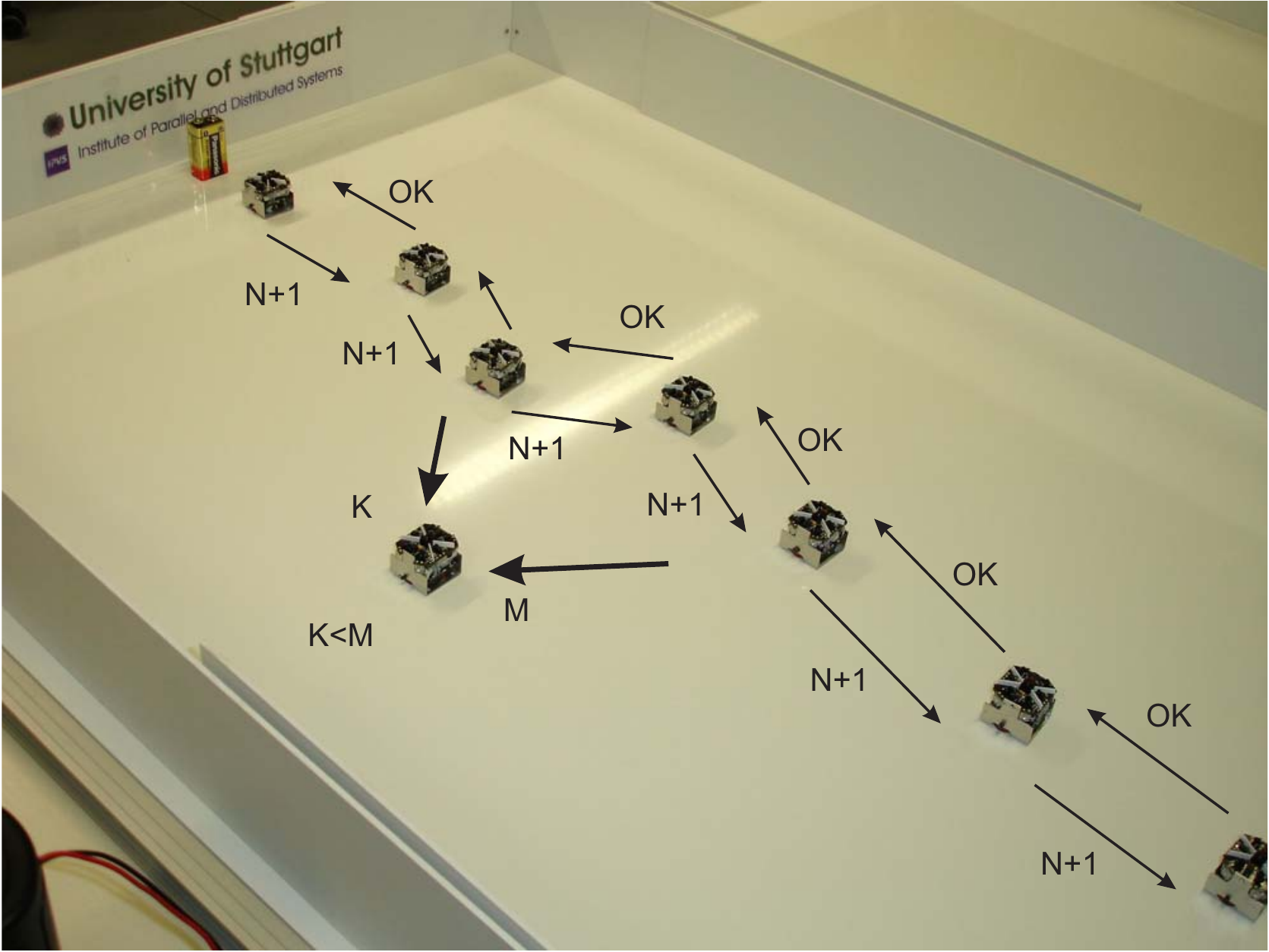}}~
\subfigure[\label{fig:comm22}]{\includegraphics[width=0.49\textwidth]{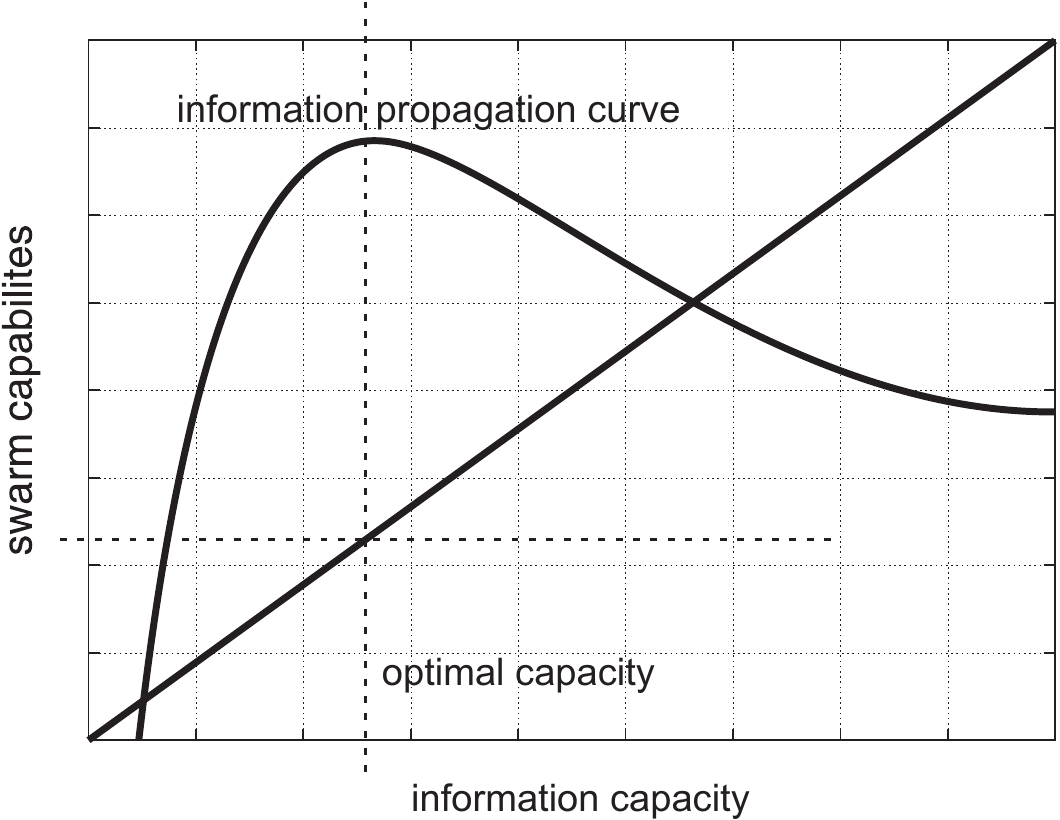}}
\caption{\small \textbf{(a)} Global collective connectivity in experiments with the creation of communication street. The messages are globally propagated during this "street". The collective behavior is regulated through circulation of these messages, e.g. the robot is navigating along the "street" based on the "gradient" of messages; \textbf{(b)} Qualitative dependency between swarm capabilities and principal factor for the case of global connectivity.
\label{fig:comm2} }
\end{figure}
As observed from Table~\ref{tab_swarm_cap}, almost no information-based and no functionality-based types of swarm behavior are possible with this mechanism. To achieve more advanced cooperation, robots should propagate their own information over a swarm. For example, a scout found some relevant resource and each robot should become aware about this event. This mechanism required that a robot, when getting a message, will propagate this message further. We denote this as a \emph{global connectivity}. It needs to take into account that this "global" mechanism is still produces by local robot-robot interactions.

The collective capabilities in this case are primarily defined by information capacity of the swarm. The more messages of different type can be propagated through the swarm, the more diverse is the resulting collective behavior. We estimate qualitatively this dependency as linear or closely-linear, see Fig.~\ref{fig:comm2}. However, the information propagation depends in turn on the information capacity (among other factors) and has a typical form shown in Fig.~\ref{fig:comm12}. {\it The collective behavior is regulated through circulated messages and swarm capabilities are primarily defined by information capacity of the swarm}. This mechanism allows a multi-functional behavior of robotic group.

\begin{figure}[ht]
\centering 
\includegraphics[width=0.9\textwidth]{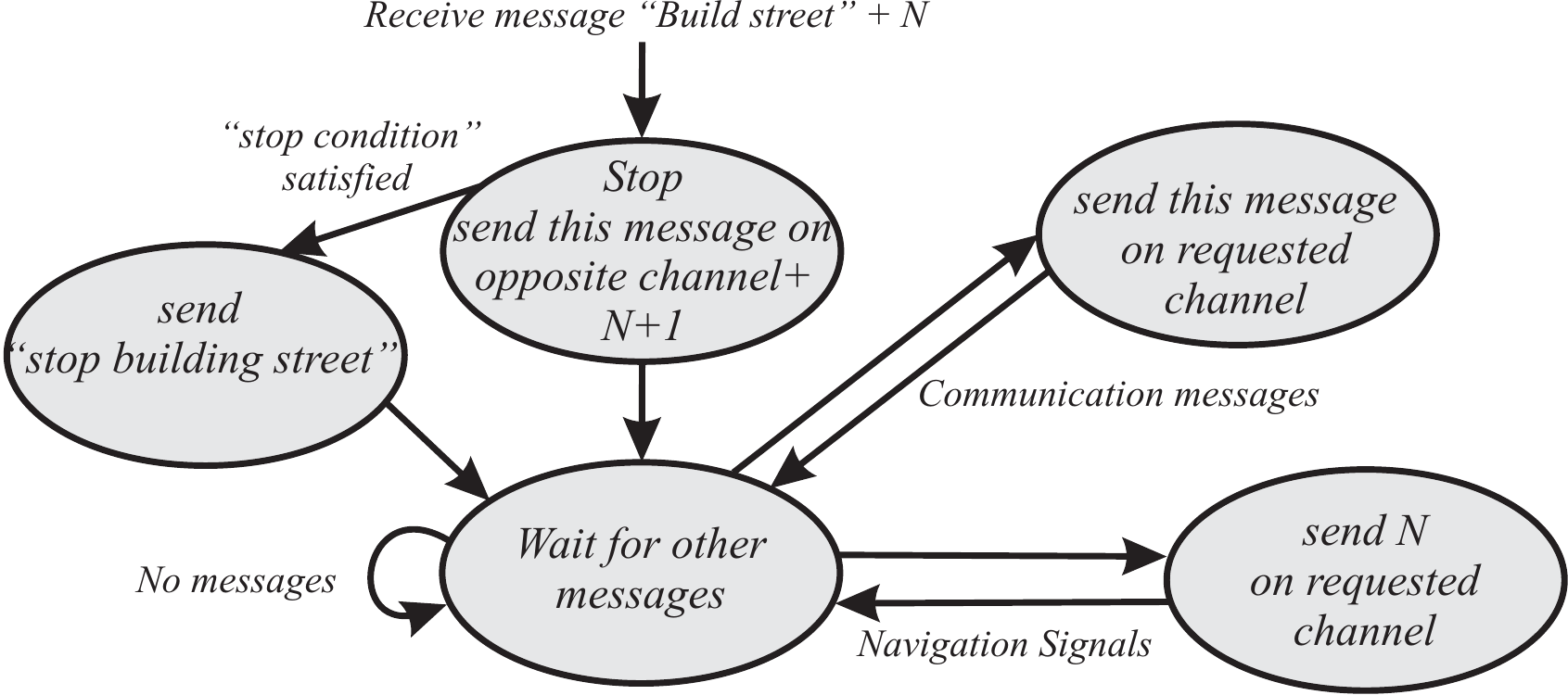}
\caption{\small Structure of the algorithm for building the communication street. \label{fig:commStrAlg} }
\end{figure}

The example of this case can be given by creating the communication street, see Fig.~\ref{fig:comm2}. The communication street is a kind of swarm's peer-to-peer network. The robots are staying within communication radii of each other. In this way, there are no clusterization effects and messages can be easily and quickly propagated in the swarm. In the experiment, shown in Fig.~\ref{fig:comm2}, robots receive some number $N$ and send $N+1$. At the end of the sequence, they return the confirmation "OK", after the cycle can be repeated. Other robots can navigate along this "street" based on the "gradient" in these signals, so they know the direction of motion. The structure of the "communication street" algorithm is shown in Fig.~\ref{fig:commStrAlg}.

\begin{figure}[ht]
\centering
\subfigure[1 sec.]{\includegraphics[width=0.33\textwidth]{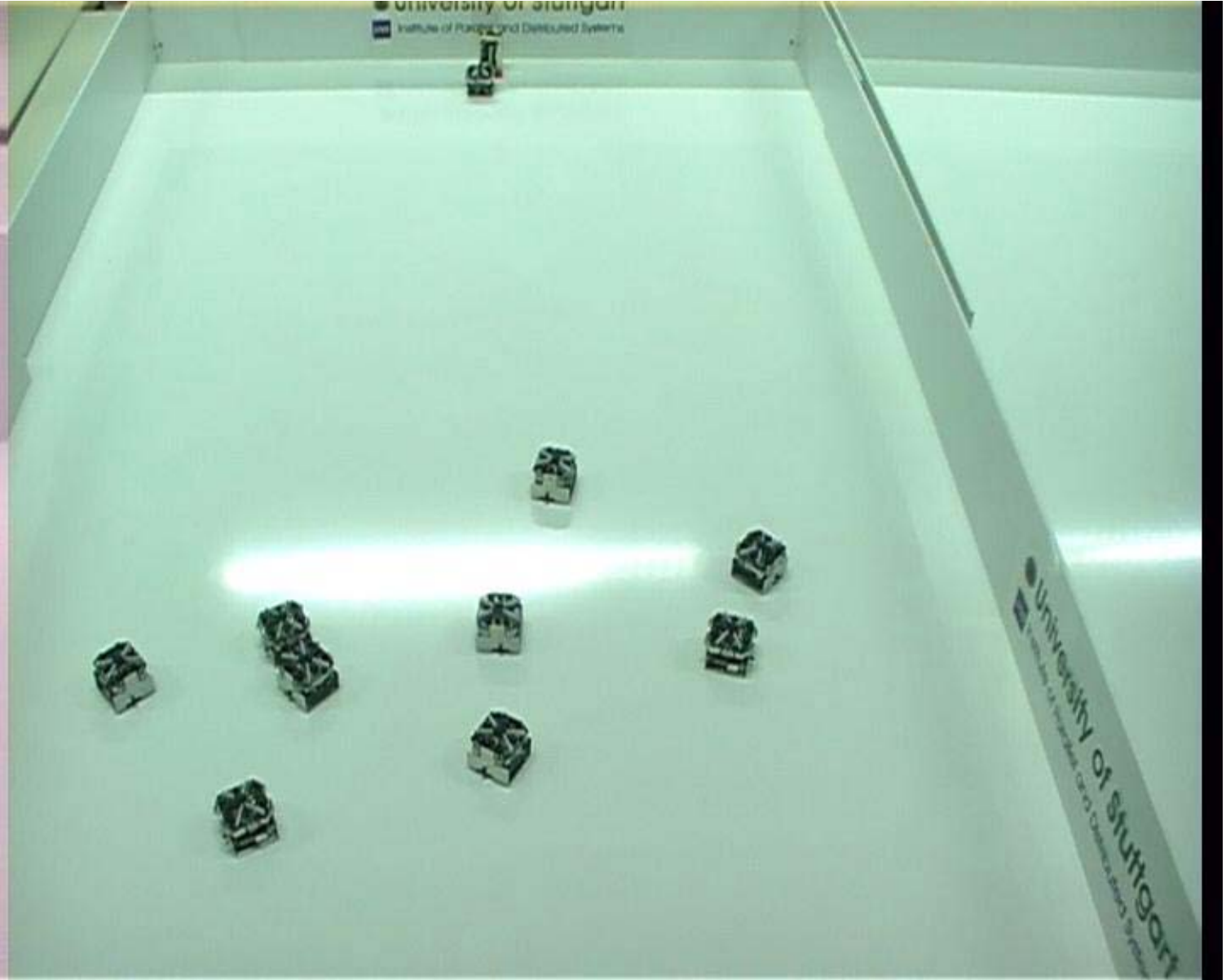}}~
\subfigure[2 sec.]{\includegraphics[width=0.33\textwidth]{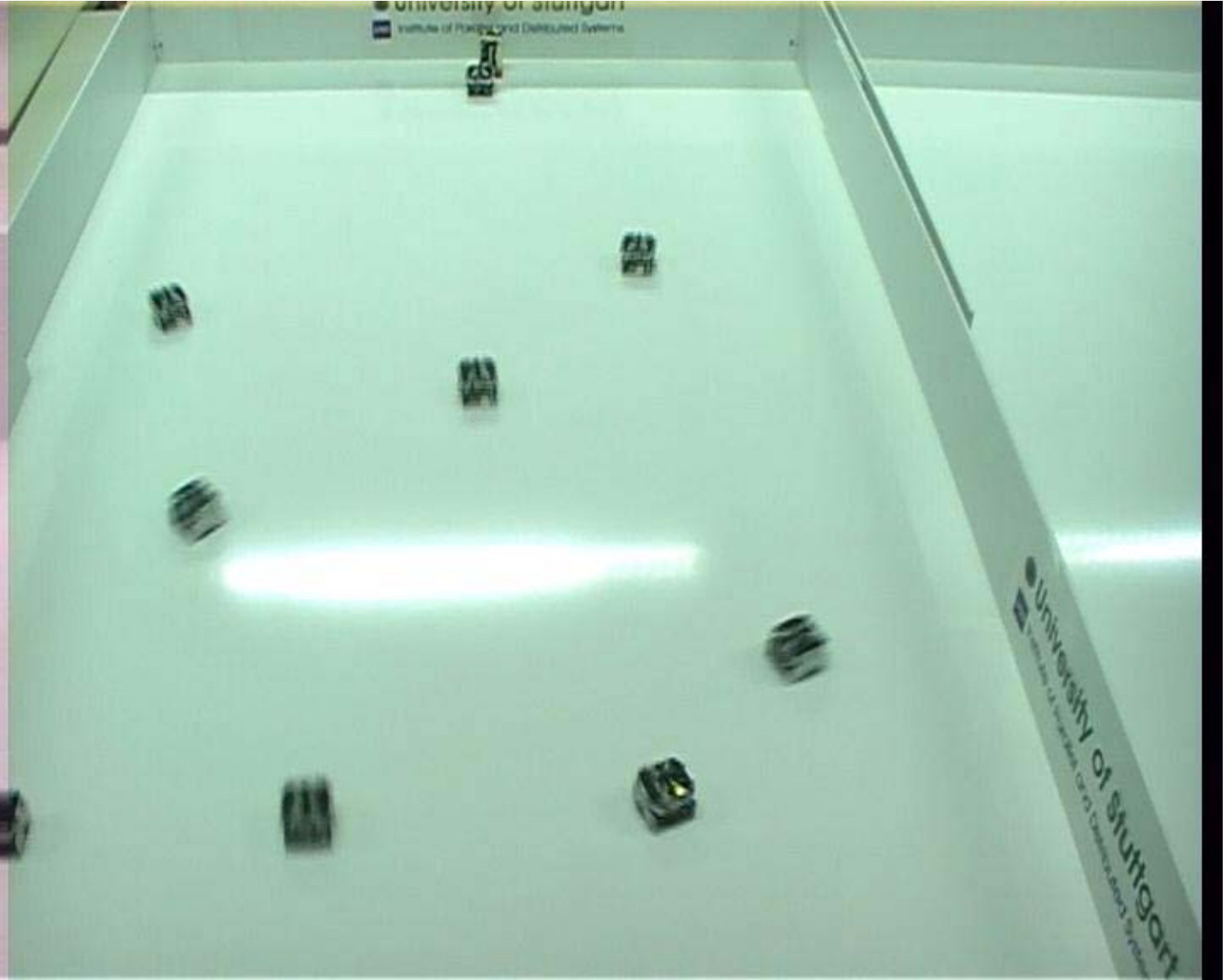}}~
\subfigure[3 sec.]{\includegraphics[width=0.33\textwidth]{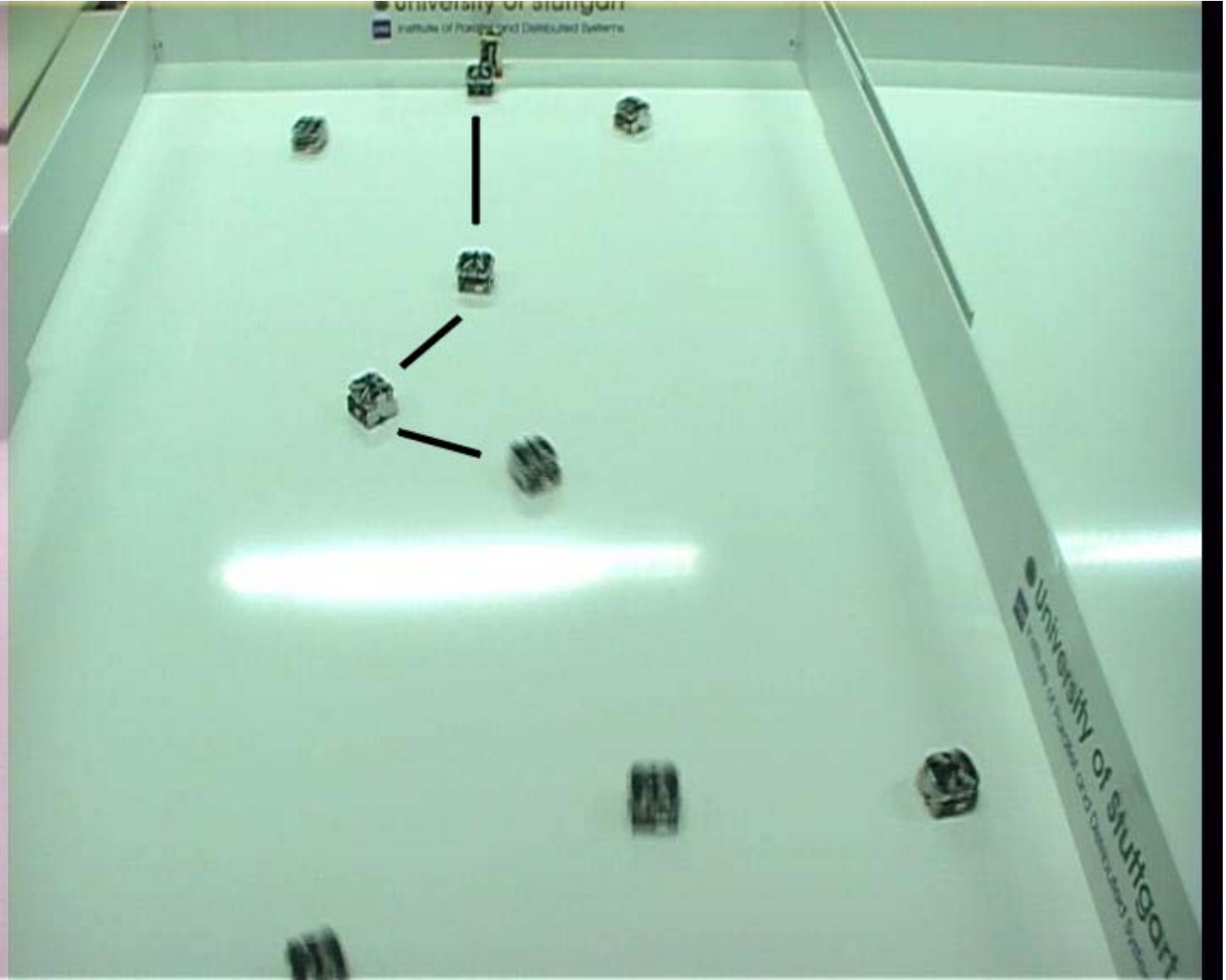}}\\
\subfigure[4 sec.]{\includegraphics[width=0.33\textwidth]{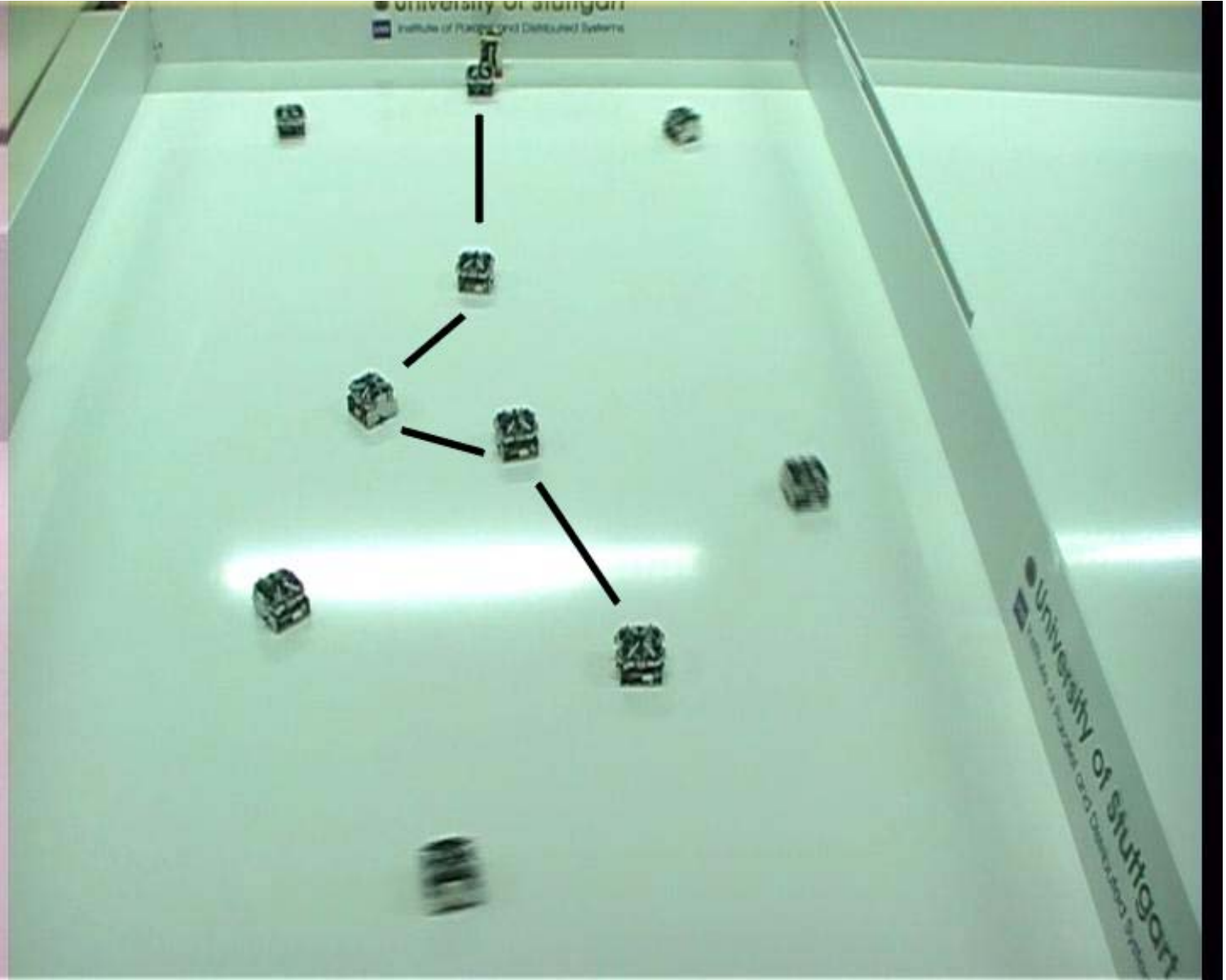}}~
\subfigure[5 sec.]{\includegraphics[width=0.33\textwidth]{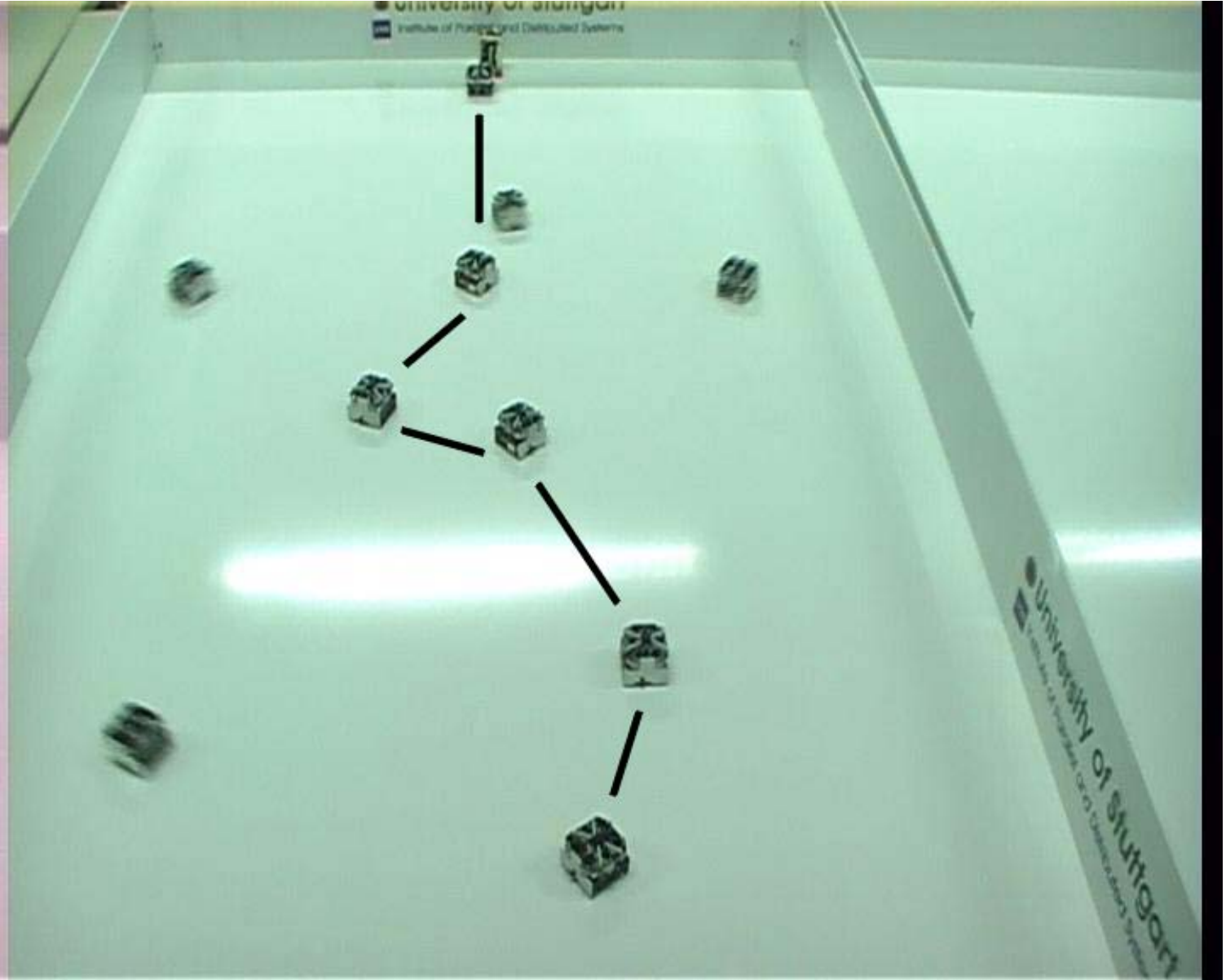}}~
\subfigure[6 sec.]{\includegraphics[width=0.33\textwidth]{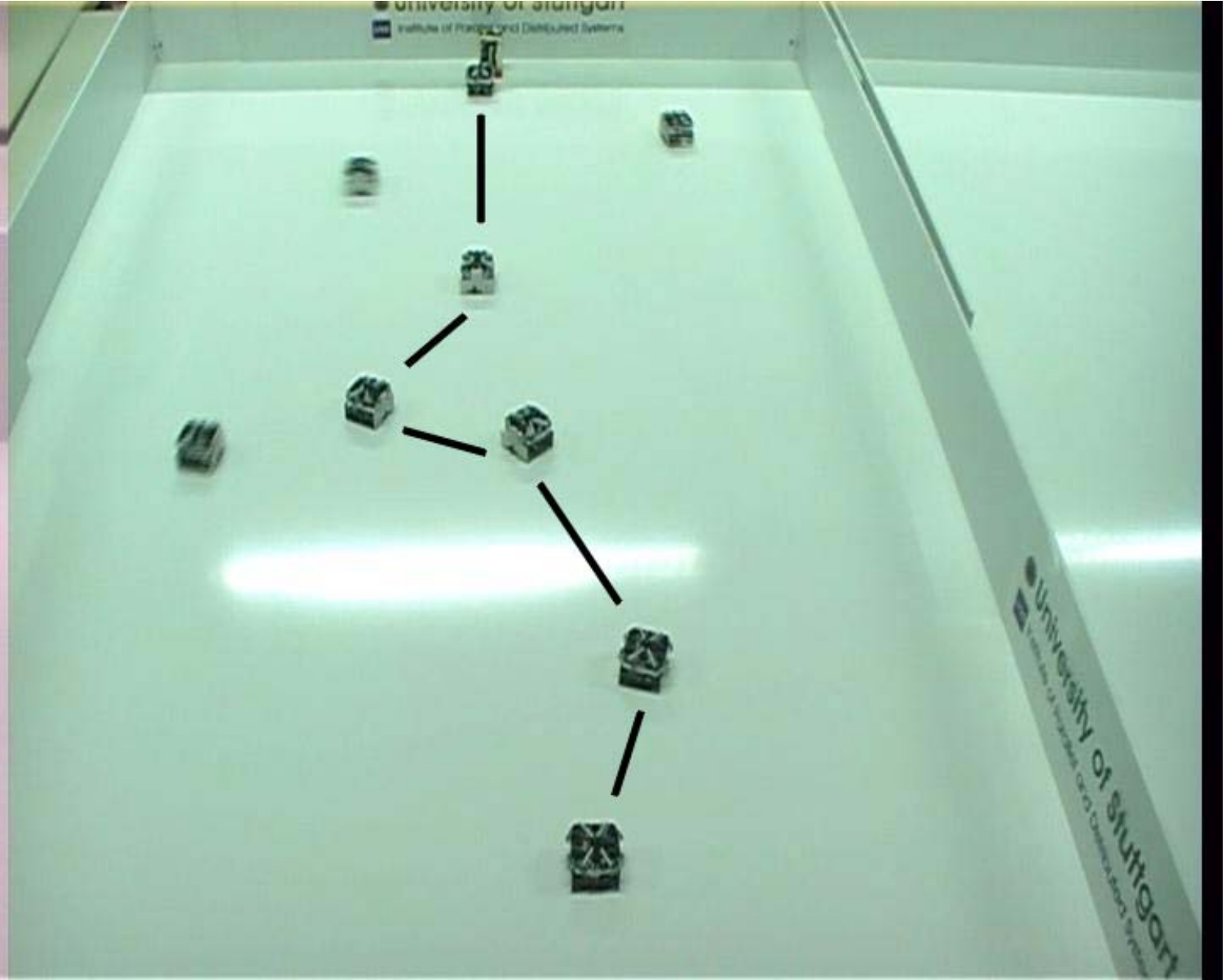}}\\
\subfigure[7 sec.]{\includegraphics[width=0.33\textwidth]{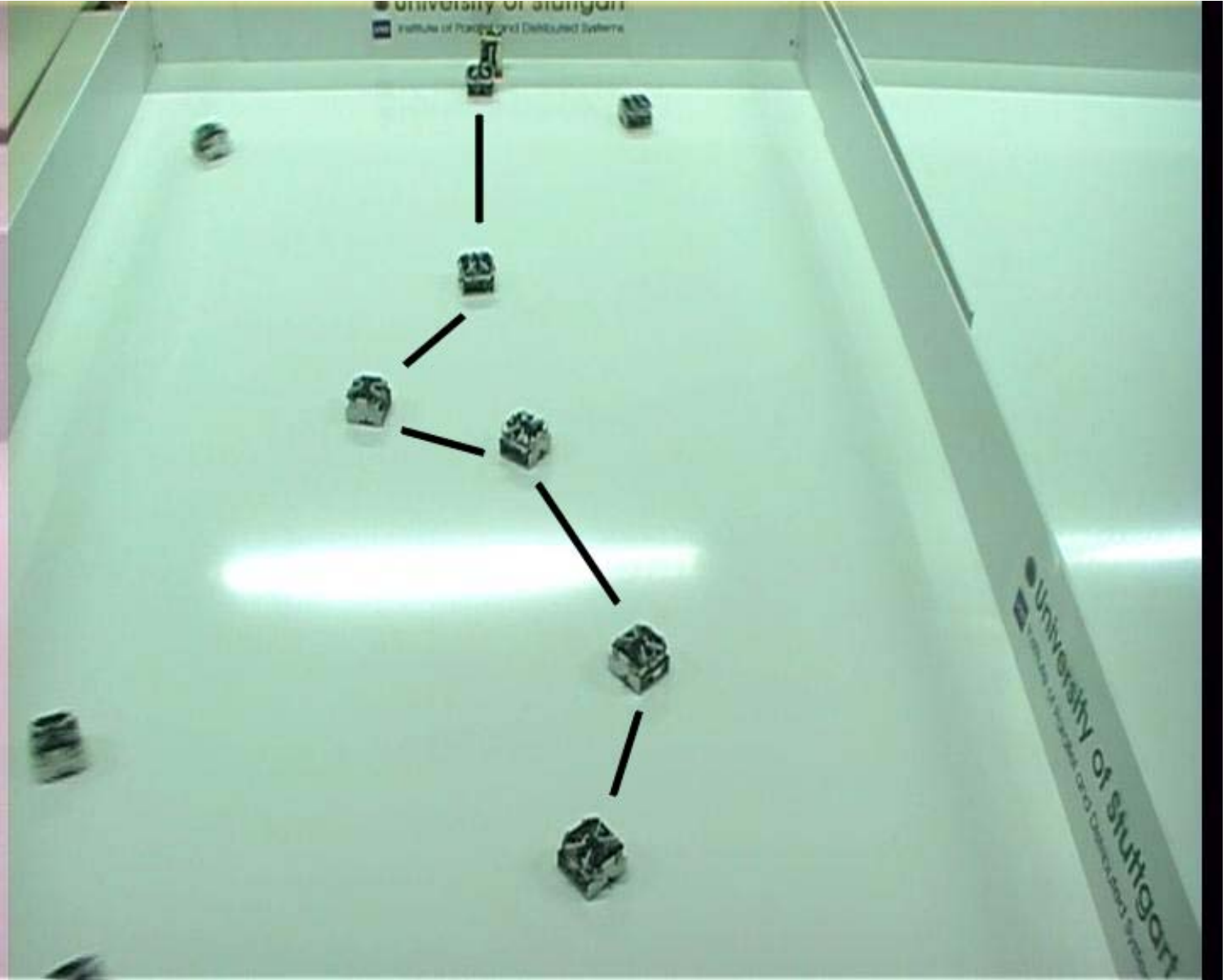}}~
\subfigure[8 sec.]{\includegraphics[width=0.33\textwidth]{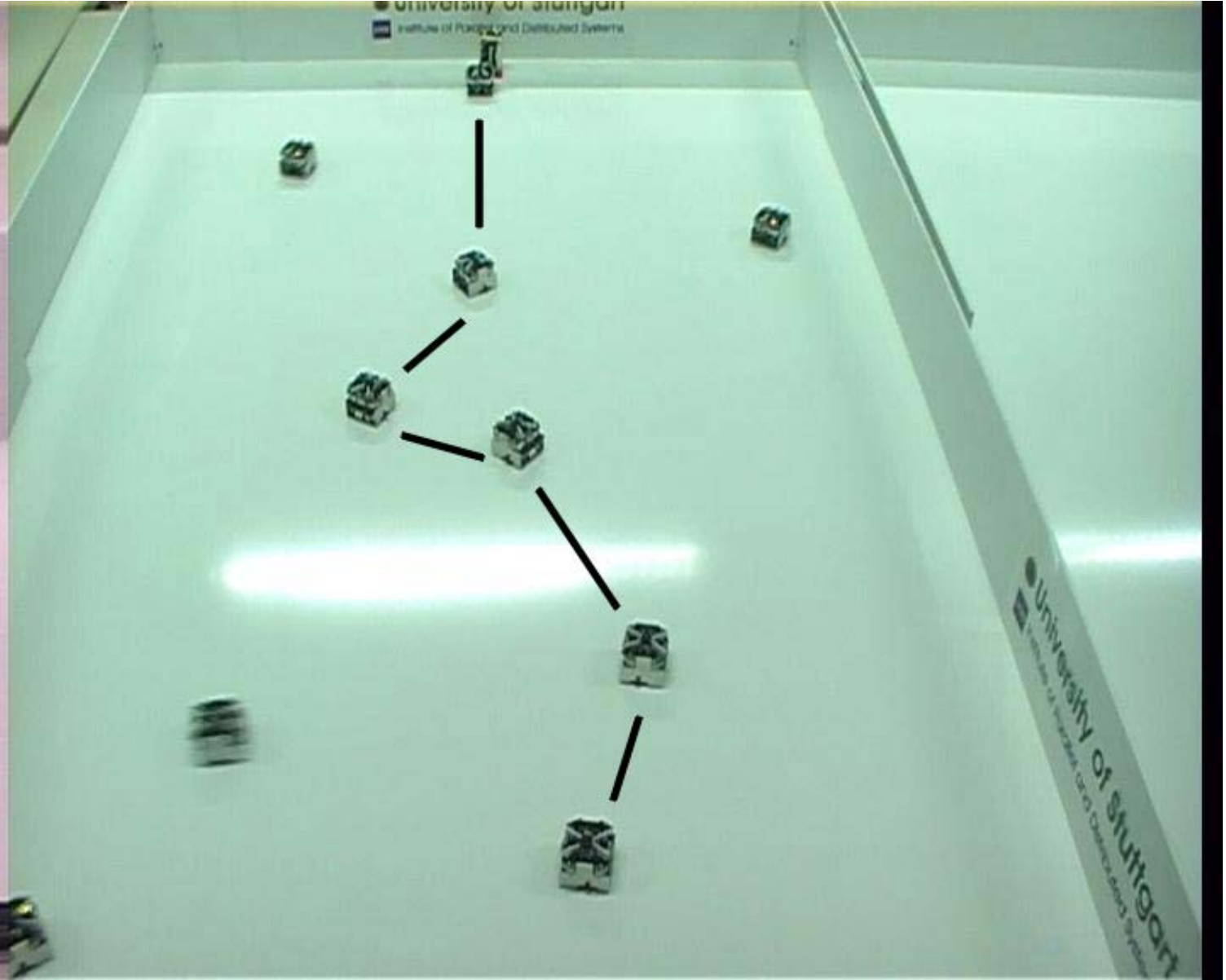}}~
\subfigure[9 sec.]{\includegraphics[width=0.33\textwidth]{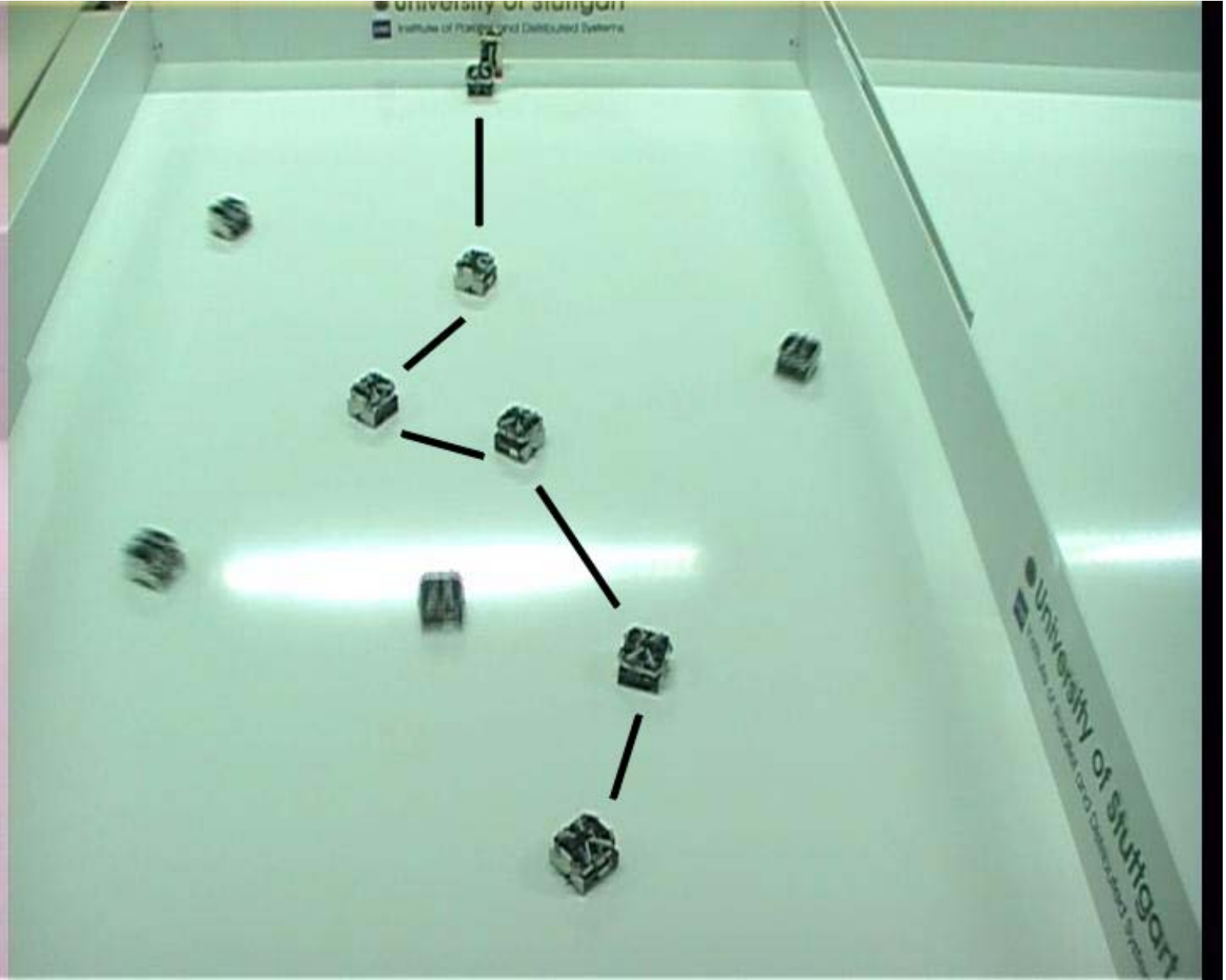}}~
\caption{\small Building the communication street. Lines show
the established connections between robots. \label{fig:commStr} }
\end{figure}

The building starts when somebody sends the first signal "build street" and $N$ on a specific channel. It can be scout or landmark robots. Usually, communication street is created between two points (two landmarks), in this case two streets are growing until they intersect. Any other robot, receiving this message, stops and checks the finishing condition. They are other streets or landmarks, or $N>threshold$. When the condition is not satisfied, it sends this message and $N+1$ further ("send" means -- send as long as another robot receives this message). When the robot is on communication street, its behavior is regulated by transferring messages, it can propagates messages on one or all IR-channels or e.g. can send "navigation signals". In Fig.~\ref{fig:commStr} we show several  images (video is available at \emph{www.swarmrobot.org}), which demonstrate building the communication street.
The whole building process take only 5 sec. for 6 robots in the street. The algorithm is well scalable at least for $<15$ robots (we cannot make more large communication streets in this arena). Propagation of information on the communication street take about 1sec. for 6 robots and about 2 sec. for $<15$ robots.

\subsection{Feedback connectivity}

\begin{figure}[htp]
\centering
\subfigure[]{\includegraphics[width=0.56\textwidth]{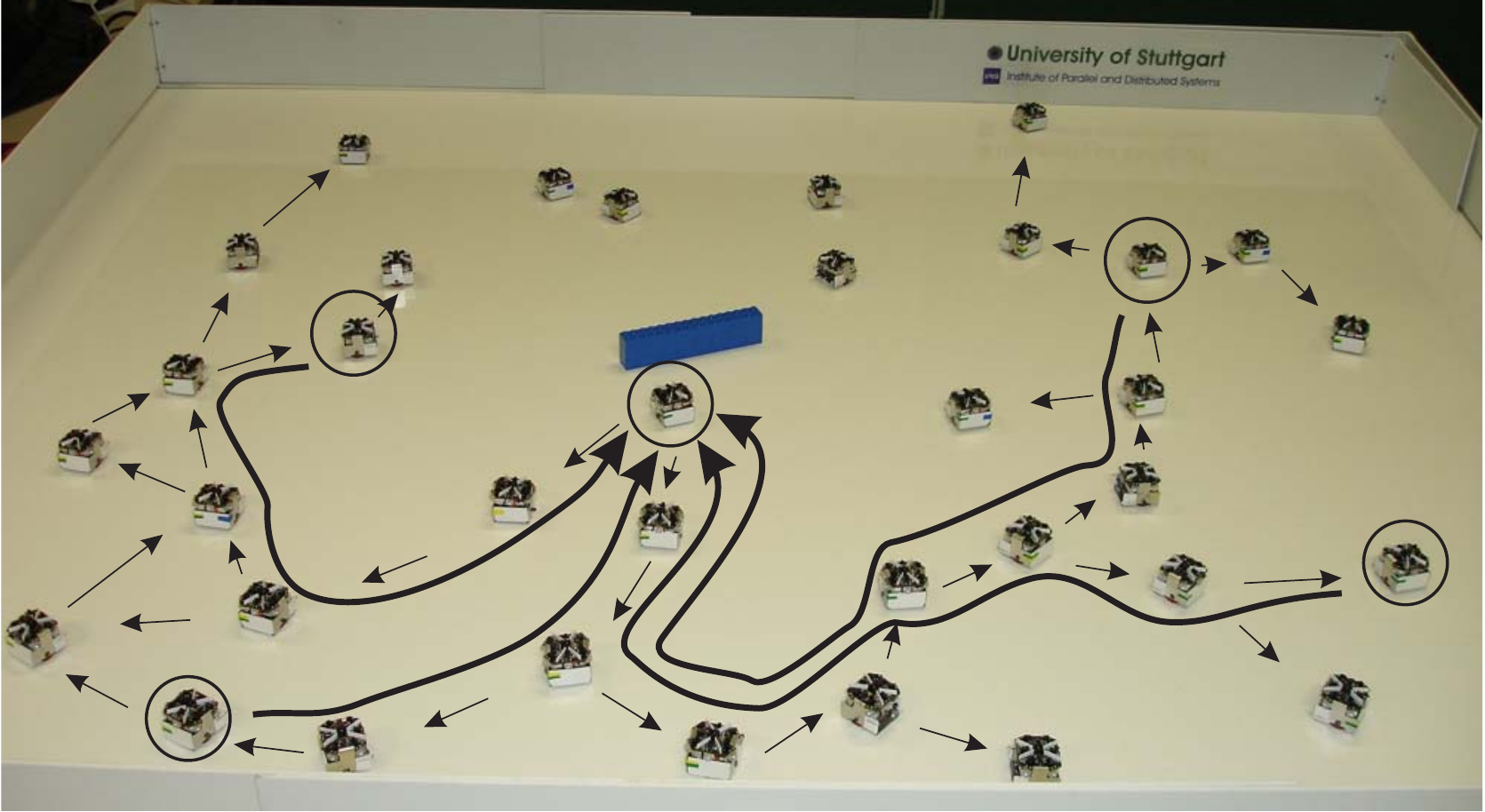}}~
\subfigure[]{\includegraphics[width=0.43\textwidth]{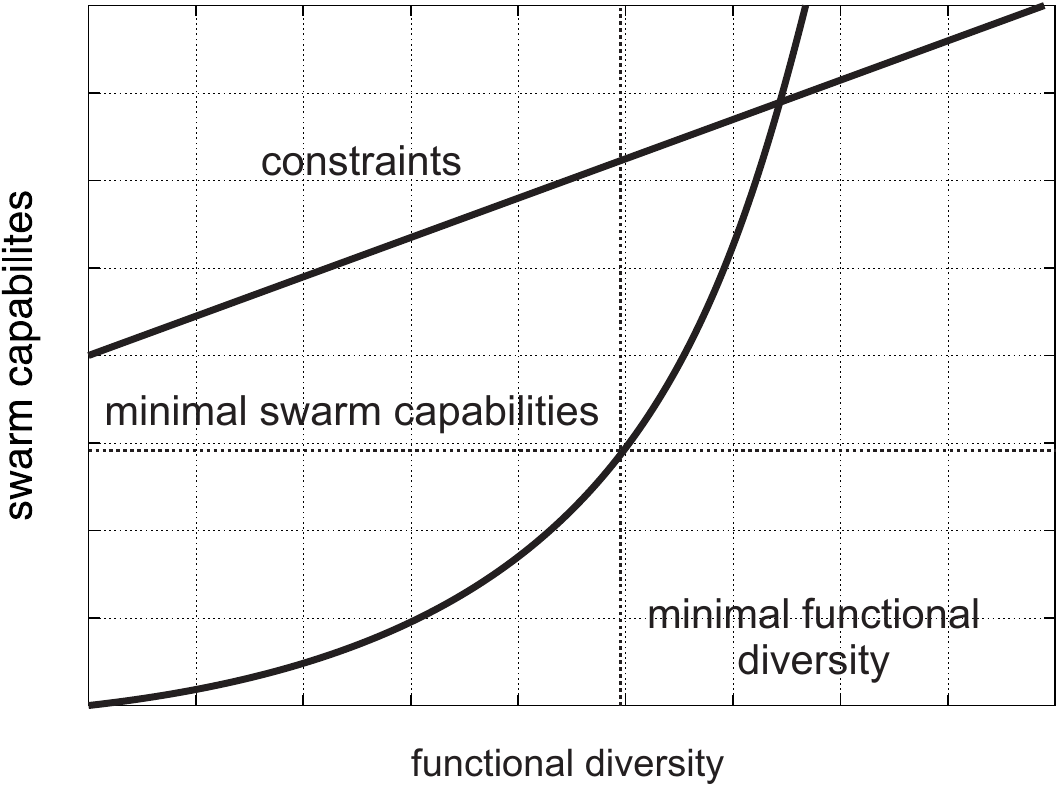}}
\caption{\small \textbf{(a)} Global feedback connectivity in experiments with the cooperative actuation. Robots, equipped with color sensors are marked by circles. Thin arrow points to global propagation of messages, thick arrow points to feedback messages. Behinds are robots that do not have any connections to the rest of the robots -- this cluster is disconnected from the swarm. \textbf{(b)} Qualitative dependency between swarm capabilities and principal factor for the case of feedback connectivity.
\label{fig:comm3} }
\end{figure}

The global collective connectivity provides messages transfer over the swarm, however this is not enough when robots have to receive the feedbacks on their own messages. It can be e.g. the request for specific resources, team building of robots with specific capabilities (robots are heterogeneous) or asking for a collective state. The communication mechanisms, when the robots are able to receive answers to their own messages, we denote the \emph{feedback connectivity}. To give an example for this case, we consider the experiment with a team building for cooperative actuation, see Fig.~\ref{fig:comm3}.
In this experiment, the robots-scout, equipped with the color sensor, found the blue object. It sends request to the swarm and ask about support -- it looks for robots with a specific functionality (in this case also equipped with the color sensor). The behavior of robot-scout (and also the swarm) depends on the feedback signals of other robots with color sensors: when there are no such robots available, scout will look further; when at least two other robots give the feedback, the scout will wait them. The mechanism of the feedback sending is relatively complex: the robots should know the sender and recipience of messages, know the terminating conditions.

We do not have now enough experimental material to estimate the primary factor influencing the collective capabilities. We assume that in the case of "individual-to-individual" cooperation in heterogeneous swarms this is defined by a functional diversity of robots. The more different types of robots are in swarm, the more different activities a swarm can demonstrate. This relation has a combinatorial character and therefore seems to be (a) exponential, and (b) limited by kinetic constraints~\cite{Kornienko_S04}.  {\it The collective behavior is regulated in this case by a cooperation between individual robots and swarm capabilities are expected to be primarily defined by a functional diversity of robots.}

\section{Conclusion}
\label{sec:conclusion}

In this paper we discussed several aspects of communication and global information transfer in a robot swarm. Three communication mechanisms are discussed, which create different coordination and scaling capabilities of swarm systems. These mechanisms are observed not only in the mentioned swarm experiments, but also in the further development of swarm robotics such as a transition between swarm and reconfigurable systems~\cite{Kornienko_S07} or evolutionary approaches running online and onboard of a robot platform~\cite{Kernbach09Platform} (see general overview of different approaches towards artificial organisms in~\cite{Levi10}).

\begin{figure}[ht]
\centering
\includegraphics[width=0.9\textwidth]{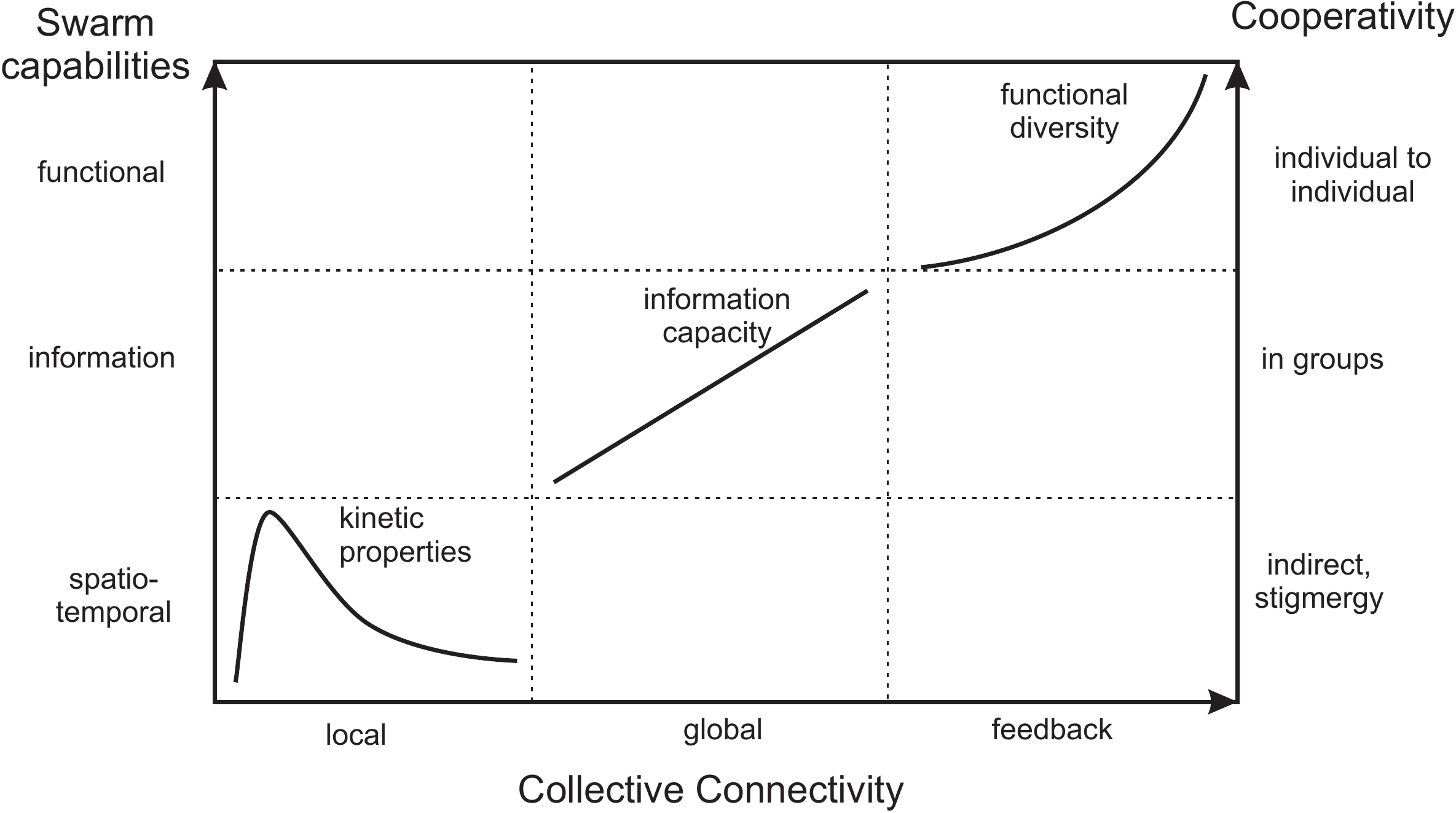}
\caption{\small Summarization of different cases of information transfer in a swarm and, as a result, different cases of cooperativity and swarm capabilities. \label{fig:comCap} }
\end{figure}

We summarized these three cases of information transfer in Fig.~\ref{fig:comCap}.
They have different character not only of underlying algorithms, but also differ in scalability, principal factors, allowed coordination mechanisms and, the most important, in the degree of common knowledge. Thus, we assume that these three cases of information transfer lead to qualitatively different swarm capabilities. 

It seems that biological collective systems, like bees or ants, use primarily the local collective connectivity. In opposite, the MAS approaches uses mainly feedback-based communication mechanisms. Implementing both in our experiments, we observe different robot behavior and different swarm properties for similar algorithms. {\it The macroscopic properties of a swarm depend not on the algorithmic implementation, but also on the embodiment of sensors/actuators and the used strategy of a global information transfer}. We did not performed any systematical characterization and quantitative experiments towards a description of swarm information, however these mechanism can be considered in the context of "swarm mathematics". Performing such experiments and their generalization represent further works, allowing better understanding the phenomenon of collective intelligence.

\small
%\bibliographystyle{unsrt}
%\bibliography{../bibl_sk,../own_bibl_sk}

\end{document}